\documentclass{article}

\usepackage{microtype}
\usepackage{graphicx}
\usepackage{subcaption}
\usepackage{booktabs}
\usepackage{tikz}
\usetikzlibrary{arrows.meta,positioning,calc}
\usepackage{hyperref}
\usepackage{amsmath}
\usepackage{amssymb}
\usepackage{mathtools}
\usepackage{amsthm}
\usepackage[capitalize,noabbrev]{cleveref}
\usepackage[preprint]{icml2026}

\theoremstyle{plain}

\theoremstyle{definition}

\theoremstyle{remark}

\begin{document}

\twocolumn[
  \icmltitle{On the Geometric Structure of Layer Updates in Deep Language Models}

  \icmlsetsymbol{equal}{*}

  \begin{icmlauthorlist}
    \icmlauthor{Jun-Sik Yoo}{ku}
  \end{icmlauthorlist}

  \icmlaffiliation{ku}{Institute of Basic Science, Korea University, 145 Anam-ro, Seoul, South Korea 02841}
  \icmlcorrespondingauthor{Jun-Sik Yoo}{junsik@korea.ac.kr}
  \icmlkeywords{Interpretability, Transformer Dynamics, Geometry of Neural Networks}

  \vskip 0.3in
]
\printAffiliationsAndNotice{}

\begin{abstract}
We study the geometric structure of layer updates in deep language models. 
Rather than analyzing what information is encoded in intermediate representations, 
we ask how representations change from one layer to the next. We show that layerwise 
updates admit a decomposition into a dominant tokenwise component and a residual that 
is not captured by restricted tokenwise function classes.

Across multiple architectures, including Transformers and state-space models, we find 
that the full layer update is almost perfectly aligned with the tokenwise component, 
while the residual exhibits substantially weaker alignment, larger angular deviation, 
and significantly lower projection onto the dominant tokenwise subspace. This indicates 
that the residual is not merely a small correction, but a geometrically distinct component 
of the transformation.

This geometric separation has functional consequences: approximation error under the 
restricted tokenwise model is strongly associated with output perturbation, with Spearman 
correlations often exceeding 0.7 and reaching up to 0.95 in larger models. Together, these 
results suggest that most layerwise updates behave like structured reparameterizations 
along a dominant direction, while functionally significant computation is concentrated 
in a geometrically distinct residual component.

Our framework provides a simple, architecture-agnostic method for probing the geometric 
and functional structure of layer updates in modern language models.
\end{abstract}

\section{Introduction}

Deep language models transform token representations across layers, yet the structure of these transformations remains poorly understood. While a large body of prior work studies what information is encoded in intermediate representations—through probing methods, representation lenses such as the Logit Lens and Tuned Lens \cite{nostalgebraist2020logitlens, belrose2023tunedlens}, and circuit-level analyses \cite{elhage2021transformercircuitsframework, olsson2022incontextlearninginductionheads}—these approaches do not directly characterize how representations change from one layer to the next.

In this work, we take a complementary perspective and study the geometry of layerwise updates. We ask: \emph{what is the structure of the transformation that maps one layer’s representation to the next?} To answer this, we introduce a decomposition of layer updates into two components: (i) a dominant tokenwise transformation that acts independently on each token, and (ii) a residual that captures the remaining component not explained by this restricted function class.

We operationalize this decomposition by approximating each layer transition using input-conditioned tokenwise maps and analyzing the residual. Across multiple architectures, including Transformers \cite{biderman2023pythia} and state-space models \cite{gu2024mambalineartimesequencemodeling}, we find that the full layer update is almost perfectly aligned with the tokenwise component. In contrast, the residual exhibits substantially weaker alignment, larger angular deviation, and significantly lower projection onto the dominant tokenwise subspace.

This geometric separation has clear functional consequences. We observe a strong monotonic relationship between approximation error and output perturbation: transitions that are well captured by tokenwise maps tend to preserve model predictions, while poorly approximated transitions induce larger output changes, with Spearman correlations often exceeding 0.7 and reaching up to $\sim$0.95 in larger models. These results indicate that the residual is not merely a small correction, but a geometrically distinct component that is disproportionately associated with changes in model behavior.

Together, our findings suggest a structured view of layerwise dynamics: most updates behave like reparameterizations along a dominant tokenwise direction, while functionally significant computation is concentrated in a geometrically distinct residual component. Importantly, this separation is not imposed by architectural design, but emerges from a functional decomposition under restricted function classes.

Our framework provides a simple, architecture-agnostic method for analyzing the geometric and functional structure of layer updates, and offers a new perspective on how computation is organized across layers in modern sequence models.

\paragraph{Contributions.}
Our main contributions are as follows:
\begin{itemize}
\item We introduce a functional decomposition of layer updates in deep language models into a dominant tokenwise component and a residual defined under restricted function classes.

\item We show that this decomposition induces a strong geometric separation: the full update is highly aligned with the tokenwise component, while the residual exhibits substantially weaker alignment, larger angular deviation, and lower projection onto the dominant tokenwise subspace.

\item We demonstrate that this geometric structure has functional consequences: approximation error under the tokenwise model is strongly associated with output perturbation across models and layers.

\item We validate these findings across multiple architectures, including both Transformer-based and state-space language models, providing an architecture-agnostic perspective on layerwise dynamics.
\end{itemize}

\section{Related Work}

\paragraph{Representation-based interpretability.}
A central goal of interpretability research is to identify which components of neural representations correspond to functionally meaningful computation. A large body of prior work has approached this problem through \emph{probing methods}, which train auxiliary classifiers to recover linguistic or semantic properties from intermediate representations. While widely used, it is now well understood that probing performance alone does not guarantee that the recovered information is actually utilized by the model. In particular, probe accuracy can reflect the capacity of the probe rather than the structure of the underlying representation~\citep{10.1162/coli_a_00422}, and properties that are recoverable from representations may not be relevant to the task the model performs~\citep{ravichander-etal-2021-probing}. Related concerns have been raised regarding classifier-based probes, which can introduce additional confounds~\citep{zhou-srikumar-2021-directprobe}. More broadly, interpretability claims are often underspecified, leading to ambiguity in what is being measured or explained~\citep{lipton2017mythosmodelinterpretability}.

\paragraph{Layerwise prediction and representation lenses.}
Beyond probing, several lines of work analyze intermediate representations through alternative lenses. Methods such as the Logit Lens and Tuned Lens~\cite{nostalgebraist2020logitlens, belrose2023tunedlens} study how hidden states relate to output predictions, suggesting that intermediate representations often support meaningful outputs. These approaches highlight that representations may already contain predictive information, but do not explicitly characterize how transformations between layers are structured.

\paragraph{Mechanistic interpretability.}
Recent work emphasizes causal interventions on internal activations to understand model behavior, including circuit-level analyses of transformer models~\cite{elhage2021transformercircuitsframework} and studies of in-context learning mechanisms~\cite{olsson2022incontextlearninginductionheads}. Techniques such as activation patching and causal tracing measure how perturbations affect outputs. While these approaches provide valuable insights into functional components, they typically focus on localized interventions rather than providing a global description of layerwise transformations.

\paragraph{From representations to transformations.}
Taken together, existing approaches focus either on static properties of representations or on localized interventions, and therefore do not directly characterize the structure of transformations between layers. This leaves a fundamental ambiguity: observed changes in representations may arise either from \emph{coordinate reparameterizations} or from \emph{functionally meaningful updates}, and existing methods generally do not distinguish between these possibilities.

\paragraph{Geometry and architecture-agnostic perspectives.}
Recent work highlights geometric structure and gauge-like freedoms in representation spaces~\cite{PhysRevResearch.7.023005}, suggesting that multiple equivalent parameterizations may exist. Our framework is compatible with this perspective, but differs in that it provides an empirical and behaviorally grounded approach to analyzing transformations under restricted function classes across architectures such as Transformers~\cite{biderman2023pythia} and state-space models~\cite{gu2024mambalineartimesequencemodeling}.

\section{Local Reparameterizations and Structured Residuals}
\label{sec:local_reparam}

We study layerwise transformations in deep sequence models through a structured decomposition into token-local and non-local components, with a particular focus on their geometric relationship.

Let $h_l = (x_1, \dots, x_L)$ denote the hidden representation at layer $l$, where $x_i \in \mathbb{R}^d$ is the embedding of token $i$. We decompose the layer transition as
\begin{equation}
\label{eq:decomposition}
h_{l+1} = T(h_l) + r(h_l),
\end{equation}
where $T$ is a tokenwise transformation and $r$ is a residual capturing the remaining component.

\subsection{Input-Conditioned Tokenwise Transformations}
\label{subsec:tokenwise}

We define $T$ as a family of \emph{input-conditioned tokenwise maps}. For each token representation $x_i$, we consider
\begin{equation}
\label{eq:tokenwise_map}
T(x_i) = A(x_i)\, x_i,
\end{equation}
where $A(x_i) \in \mathbb{R}^{d \times d}$ depends on the input representation $x_i$.

This is not a single global linear transformation, but a family of local linear maps whose parameters vary with the input. As a result, the overall mapping is \emph{globally nonlinear} despite being locally linear at each point.

In practice, we impose only weak locality constraints (e.g., via nearest neighbors in representation space), ensuring that $A(x_i)$ varies smoothly with $x_i$.

\subsection{Restricted Function Class as a Structural Probe}
\label{subsec:restricted_class}

Our objective is not to maximize approximation accuracy, but to impose a structured function class that enables an interpretable decomposition of layerwise updates.

The class of input-conditioned tokenwise transformations is highly expressive: it captures a wide range of token-local updates, including input-dependent rescalings and rotations, without requiring cross-token interaction. At the same time, this restriction prevents the model from absorbing all variation into $T$.

This design makes the residual
\begin{equation}
\label{eq:residual}
r(h_l) = h_{l+1} - T(h_l)
\end{equation}
informative. Rather than treating approximation error as noise, we interpret it as a signal of structure that lies outside the chosen function class.

\subsection{Geometric Interpretation of the Residual}
\label{subsec:residual_interpretation}

Beyond its functional role, the residual admits a geometric interpretation. The tokenwise component $T(h_l)$ defines a \emph{dominant update direction} at each token, capturing the primary axis along which representations are transformed under the restricted function class.

The residual $r(h_l)$ then represents the component of the update that is not aligned with this dominant direction. As we will show empirically, this component is not simply a small correction, but is often weakly aligned or substantially deviated from the tokenwise update direction.

This perspective emphasizes that the decomposition in Eq.~\eqref{eq:decomposition} is not merely functional but also geometric: it separates updates into a dominant tokenwise direction and a complementary component that lies outside it.

While residual components may include contributions from cross-token interaction mechanisms such as attention or state-space mixing, we do not assume a strict correspondence. Instead, we treat the residual as an empirical probe of structure not captured by tokenwise transformations.

This geometric separation is consistent with the interpretation that the fitted tokenwise map provides a local linear approximation to the dominant tokenwise update structure at each point. In this view, non-local interactions induce deviations that lie largely outside this locally defined tokenwise direction or subspace.

\subsection{Functional Evaluation via Intervention}
\label{subsec:intervention}

To assess the functional significance of the decomposition, we perform an intervention by replacing the original transition with $T(h_l)$ and propagating the modified representation through the remaining layers.

We then measure the resulting change in the model's output distribution. This allows us to evaluate whether components captured by the tokenwise transformation preserve model behavior, and whether components outside this class are associated with larger functional changes.

\begin{figure}[t]
\centering
\includegraphics[width=0.44\textwidth]{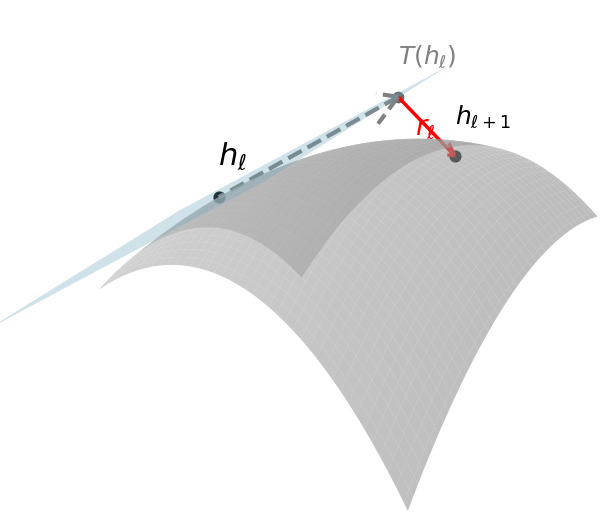}
\caption{
Illustration of the decomposition into a local linear approximation and a residual term.
The tokenwise prediction captures the dominant update direction, while the residual reflects deviations from this direction that arise from non-local structure.
}
\label{fig:framework}
\end{figure}

\paragraph{Instantiation.}
We now describe how this framework is instantiated in practice, including the choice of function classes, local fitting procedure, and evaluation metrics.

\section{Instantiation and Evaluation}
\label{sec:method}

We instantiate the framework introduced in Section~\ref{sec:local_reparam} by specifying concrete tokenwise function classes, a local fitting procedure, and evaluation metrics.
\subsection{Function Classes}

We consider several families of tokenwise transformations:
\begin{itemize}
\item Diagonal positive semi-definite (Diag-PSD) maps,
\item Low-rank linear maps,
\item Orthogonal (unitary) transformations,
\item We also consider shallow nonlinear maps (e.g., small MLPs) as an additional function class.
\end{itemize}

All transformations act independently on each token representation.

\subsection{Implementation Details}

For each anchor representation $h_\ell^{(i)}$, we construct a local neighborhood
$\mathcal{N}_k(i)$ using $k$-nearest neighbors in representation space.

We then fit a local tokenwise transformation $T_i \in \mathcal{F}$ by minimizing
the reconstruction error over this neighborhood:
\begin{equation}
T_i = \arg\min_{T \in \mathcal{F}} \sum_{j \in \mathcal{N}_k(i)}
\| h_{\ell+1}^{(j)} - T(h_\ell^{(j)}) \|^2,
\end{equation}
where $\mathcal{F}$ denotes the chosen function class (e.g., diagonal, low-rank, or MLP).

The resulting map $T_i$ is interpreted as a local approximation of the layer transition
around $h_\ell^{(i)}$, and is evaluated at the anchor to produce $T_i(h_\ell^{(i)})$.
This procedure defines a locally adaptive approximation of the layer transition,
without introducing cross-token interactions.

Because the neighborhood $\mathcal{N}_k(i)$ depends on discrete nearest-neighbor assignments,
the fitted maps may vary discontinuously across representation space. In practice, this effect
is mitigated by the high density of representations and by averaging over multiple anchors.

\paragraph{Interpolation across anchors.}
In addition to per-anchor fitting, we consider interpolating between nearby anchors
to obtain a smoother approximation. Given a query representation $h_\ell$, we identify
its nearest anchors $\{h_\ell^{(i)}\}$ and combine their corresponding maps $\{T_i\}$
using distance-based weights. The resulting interpolated transformation provides
a continuous approximation across representation space while preserving locality.

Taken together, this defines a piecewise-smooth approximation of the layer transition,
parameterized by anchor locations and local neighborhoods.

\subsection{Evaluation Metrics and Geometric Analysis}

We evaluate each fitted transformation using both functional and geometric metrics.

\paragraph{Representation Error.}
Relative $\ell_2$ error between $T(h_\ell)$ and $h_{\ell+1}$.

\paragraph{Output Perturbation.}
KL divergence between the original output distribution and the distribution after intervention.

\paragraph{Error--Perturbation Correlation.}
Spearman correlation between representation error and output perturbation, capturing the relationship between approximation quality and functional impact.

\paragraph{Directional Alignment.}
To characterize the geometric structure of the decomposition, we compare three update vectors:
\[
\Delta_{\text{full}} = h_{\ell+1} - h_\ell, \quad
\Delta_{\text{tok}} = T(h_\ell) - h_\ell, \quad
r = h_{\ell+1} - T(h_\ell).
\]

We measure alignment using absolute cosine similarity:
\[
\mathrm{Align}(v, u) = \frac{|v^\top u|}{\|v\|\|u\|}.
\]

We report alignment between:
\begin{itemize}
\item the full update $\Delta_{\text{full}}$ and the tokenwise update $\Delta_{\text{tok}}$,
\item the residual $r$ and the tokenwise update $\Delta_{\text{tok}}$.
\end{itemize}

In addition, we compute the angle between vectors to quantify angular deviation.

\paragraph{Subspace Projection.}
To assess how updates align with the dominant tokenwise structure, we compute the projection of each update onto the leading singular vectors of the local map $A(x)$.

Given the top-$k$ left singular vectors $U_k$, we measure the fraction of energy captured:
\[
\frac{\| U_k^\top v \|^2}{\|v\|^2}.
\]

We report this quantity for the full update, tokenwise update, and residual.

\paragraph{Signed Alignment (Auxiliary).}
We additionally report signed cosine similarity to detect systematic directional bias, although all primary alignment results use absolute cosine.

\section{Experiments}
\label{sec:experiments}

We empirically evaluate our framework across multiple model architectures, layers, and function classes. Our goal is to quantify how much of layer-to-layer transformation can be explained by restricted tokenwise mappings, and to understand the geometric and functional structure of the remaining residual.

\begin{figure*}[t]
  \centering

  \begin{subfigure}[t]{0.24\textwidth}
    \centering
    \includegraphics[width=\linewidth]{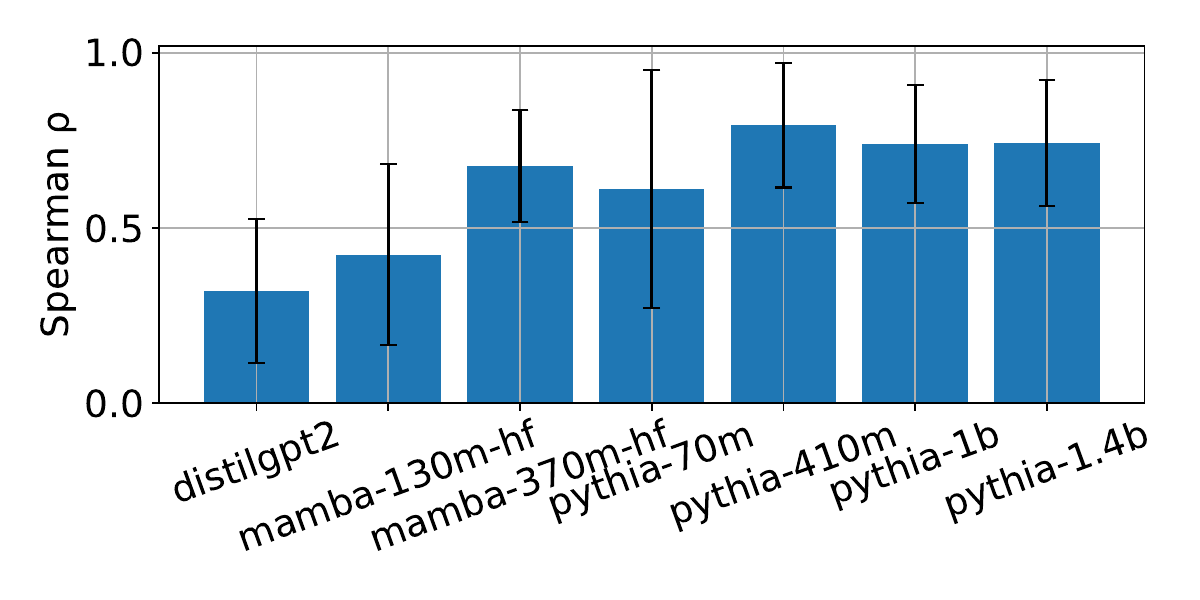}
    \caption{}
    \label{fig:residual_bar}
  \end{subfigure}
  \hfill
  \begin{subfigure}[t]{0.24\textwidth}
    \centering
    \includegraphics[width=\linewidth]{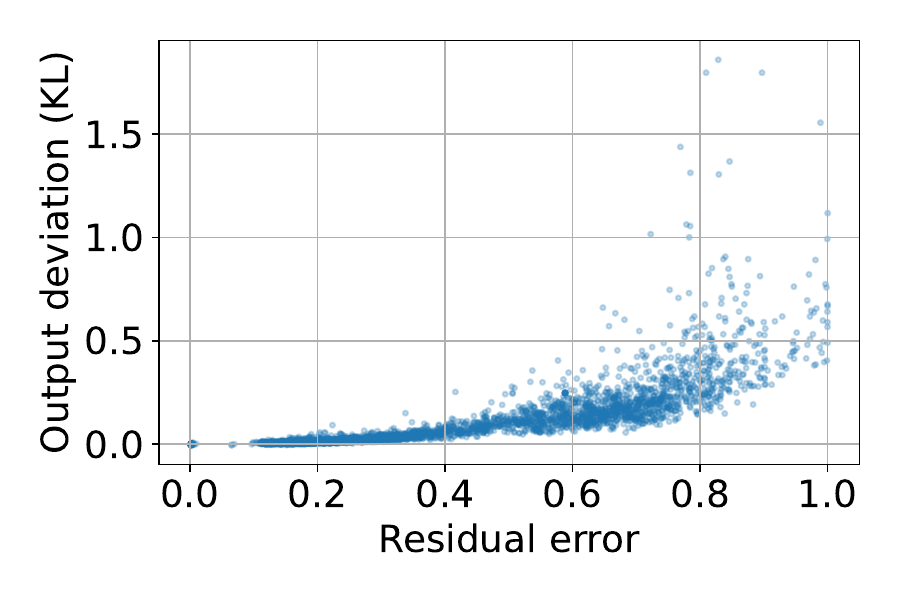}
    \caption{}
    \label{fig:residual_scatter}
  \end{subfigure}
  \hfill
  \begin{subfigure}[t]{0.24\textwidth}
    \centering
    \includegraphics[width=\linewidth]{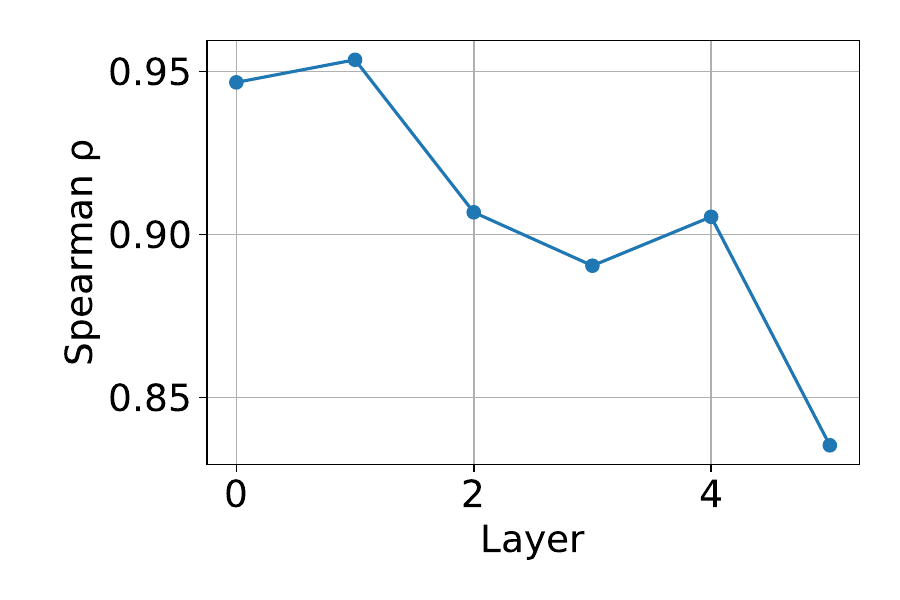}
    \caption{}
    \label{fig:residual_layer_align}
  \end{subfigure}
  \begin{subfigure}[t]{0.24\textwidth}
    \centering
    \includegraphics[width=\linewidth]{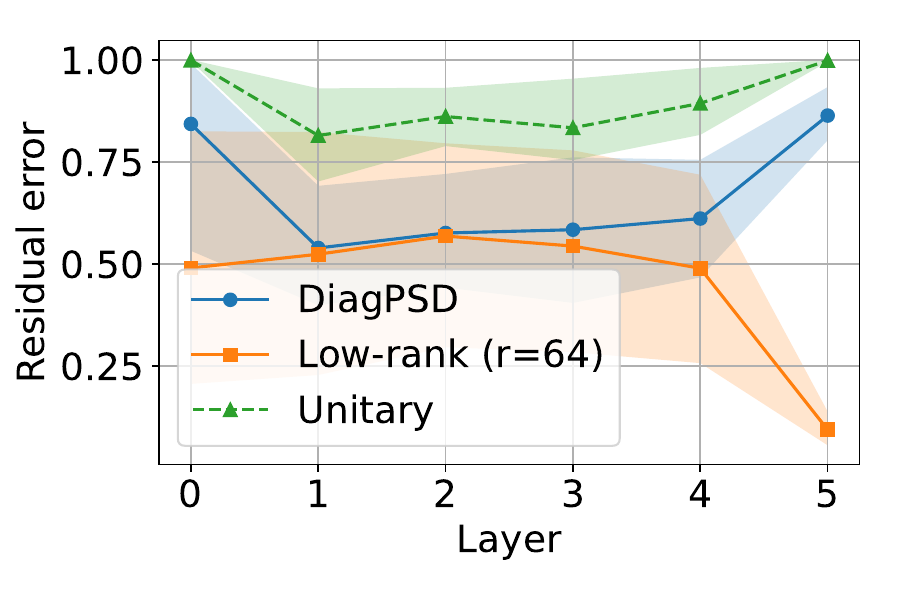}
    \caption{}
    \label{fig:residual_layer_mag}
  \end{subfigure}

  \caption{
  \textbf{Residual-output relationship across models and layers.}
  (a) Across architectures, residual error strongly correlates with output deviation.
  (b) Representative token-level scatter (Pythia-1B).
  (c) Layer-wise variation of residual--output alignment.
  (d) residual magnitude varies across layers, revealing structured regimes.
  }
  \label{fig:residual_main}
\end{figure*}

\subsection{Models and Setup}

We consider a diverse set of pretrained language models, including \textbf{DistilGPT2}, \textbf{Pythia-70M}, \textbf{Pythia-410M}, \textbf{Pythia-1B}, \textbf{Pythia-1.4B}, \textbf{Mamba-130M}, and \textbf{Mamba-370M}. This selection spans Transformer-based architectures and state-space sequence models without explicit attention mechanisms.

For all models, we extract hidden representations at successive layers and collect token-level pairs $(h_\ell, h_{\ell+1})$ using teacher-forced decoding. Experiments are conducted on WikiText with fixed sequence length, batch size, and random seeds.

\subsection{Local Tokenwise Approximations}

We approximate layer transitions using restricted tokenwise function classes: diagonal PSD maps, orthogonal transformations, and low-rank linear maps. All transformations act independently on each token representation.

For each function class, we fit $T$ locally using $k$-nearest neighbors in representation space, minimizing
\[
\| h_{\ell+1} - T(h_\ell) \|^2.
\]
We report the resulting relative error (RelErr) as a measure of approximation quality.

\subsection{Geometric Structure of Layer Updates}
\begin{figure}[t]
\centering
\begin{subfigure}[t]{0.59\linewidth}
    \centering
    \includegraphics[width=\linewidth]{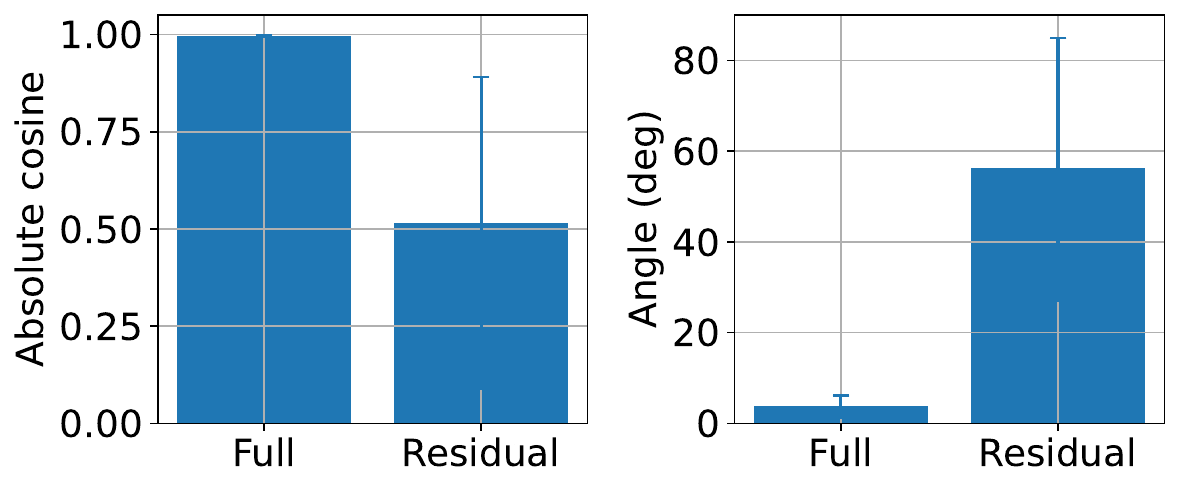}
    \caption{Alignment summary.}
    \label{fig3a:model_rho}
\end{subfigure}
\hfill
\begin{subfigure}[t]{0.39\linewidth}
    \centering
    \includegraphics[width=\linewidth]{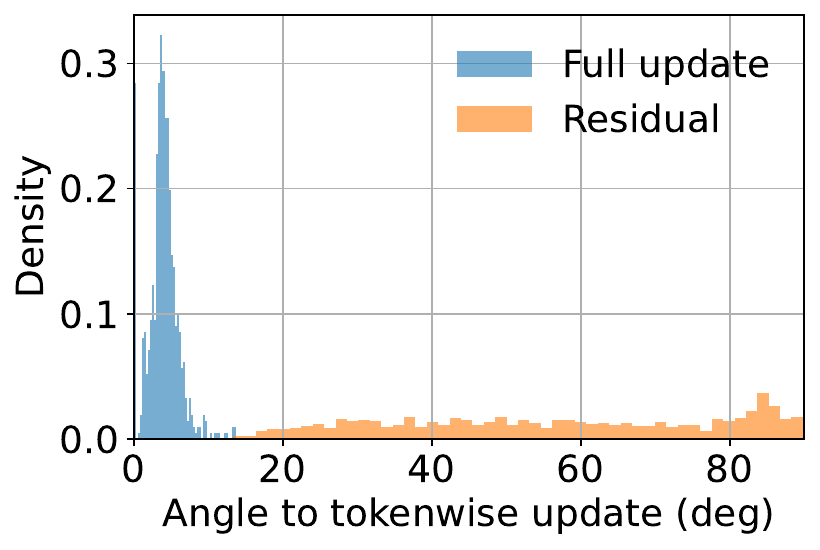}
    \caption{Angular deviation distribution.}
\end{subfigure}
\caption{
\textbf{Geometric structure of layer updates.}
The full update is strongly aligned with the tokenwise approximation (left),
while the residual exhibits large angular deviation (right), indicating a geometrically distinct component.
}
\label{fig:geom_alignment}
\end{figure}

\begin{figure}[t]
\centering
    \includegraphics[width=.99\linewidth]{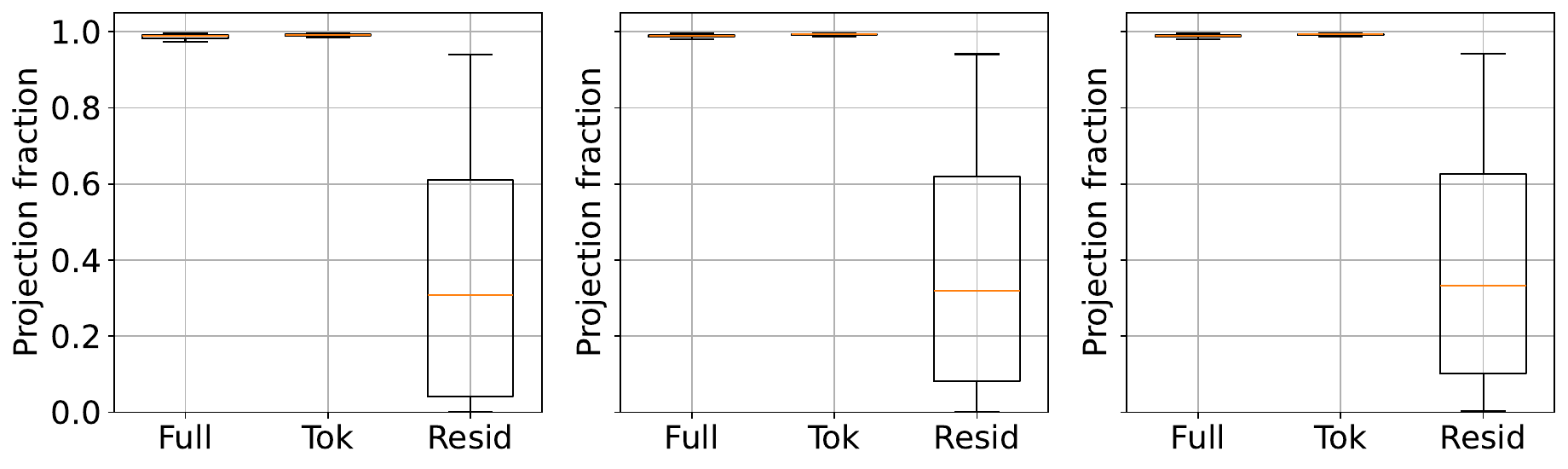}
\caption{
\textbf{Projection onto dominant tokenwise subspace.}
The full and tokenwise updates lie almost entirely within a low-dimensional subspace,
while the residual exhibits significantly lower projection, confirming its geometric separation. Results shown for top-1, top-4, and top-8 singular vectors.
}
\label{fig:geom_subspace}
\end{figure}

We analyze the geometric relationship between the full update, tokenwise approximation, and residual.

As shown in Figure~\ref{fig:geom_alignment}, the full update is almost perfectly aligned with the tokenwise approximation, with cosine similarity close to 1 and small angular deviation.

In contrast, the residual exhibits a broad angular distribution, with many tokens exceeding 60 degrees, indicating that it does not lie along the dominant direction.

Figure~\ref{fig:geom_subspace} further shows that while the full update is concentrated in a low-dimensional subspace, the residual is not, confirming that it forms a geometrically distinct component of the transformation.

\subsection{Functional Alignment: Residual vs Output}

We evaluate the functional impact of approximation error by measuring output sensitivity. Specifically, we compute the KL divergence between the model's output distribution under the original representation and under the transformed representation $T(h_\ell)$.

Across models and layers, we compute the Spearman rank correlation between RelErr and output divergence.

As shown in Figure~\ref{fig:residual_bar} and Table~\ref{tab:model_summary}, we observe consistently strong correlations, often exceeding $0.7$ in larger models.

\begin{table}[t]
\centering
\small
\begin{tabular}{lc}
\toprule
Model & Mean Spearman $\rho$ \\
\midrule
Pythia-70m & 0.906 \\
Pythia-410m & 0.837 \\
Pythia-1b & 0.804 \\
Pythia-1.4b & 0.883 \\
distilgpt2 & 0.321 \\
Mamba-130m & 0.605 \\
Mamba-370m & 0.865 \\
\bottomrule
\end{tabular}
\caption{
Mean alignment between approximation error and output perturbation across models.
}
\label{tab:model_summary}
\end{table}

At the token level (Figure~\ref{fig:residual_scatter}), we observe a clear monotonic trend: tokens with larger residual errors induce larger output changes. This indicates that approximation failure is not random, but systematically aligned with functional importance.

Table~\ref{tab:model_summary} reports the unified low-rank (r=64) results across all models and layers. Figure~\ref{fig:residual_bar} visualizes the same setting at the model level, showing consistent trends across architectures.

\subsection{Layerwise Structure and Regimes}

The strength of this alignment varies across layers. Figures~\ref{fig:residual_layer_align} and \ref{fig:residual_layer_mag} show that both residual magnitude and its functional alignment exhibit structured variation across depth.

In particular, intermediate layers often display higher residuals and weaker alignment, suggesting regimes where tokenwise approximations fail to capture key transformations.

\subsection{Function Class Comparison}

We compare linear and nonlinear function classes across regimes.

\begin{table*}[t]
\centering
\small
\begin{tabular}{lcccccc}
\toprule
Regime & RelErr (Lin) $\downarrow$ & RelErr (MLP) $\downarrow$ 
& KL (Lin) $\downarrow$ & KL (MLP) $\downarrow$
& $\rho$ (Lin) $\uparrow$ & $\rho$ (MLP) $\uparrow$ \\
\midrule
Low  & 0.18 & 0.26 & 6.1e-4 & 1.30e-3 & 0.80 & 0.62 \\
Mid  & 0.43 & 0.46 & 2.73e-3 & 1.88e-3 & 0.59 & 0.56 \\
High & 0.62 & 0.57 & 7.50e-3 & 2.53e-3 & 0.06 & 0.53 \\
\midrule
Overall 
& 0.41 & 0.43 
& 3.63e-3 & 1.91e-3 
& 0.82 & 0.73 \\
\bottomrule
\end{tabular}
\caption{
Comparison between local linear maps and small MLP baselines across regimes.
}
\label{tab:mlp_vs_linear}
\end{table*}

In low-error regimes, linear maps exhibit stronger alignment between residual magnitude and output perturbation. In high-error regimes, this alignment degrades for linear maps, while MLPs partially recover it due to increased expressivity.

This highlights a trade-off: more expressive models reduce residual magnitude but may weaken interpretability.

\subsection{Architectural Trends}

We observe systematic differences across architectures.

Smaller Transformer models (e.g., DistilGPT2) are well approximated by diagonal transformations, indicating simpler tokenwise structure. Larger models (e.g., Pythia) benefit significantly from low-rank maps, suggesting richer but still structured tokenwise dynamics.

Mamba models exhibit a similar decomposition pattern, despite lacking attention, indicating that the observed structure is not specific to attention-based architectures.

\subsection{Summary}
Across architectures and layers, we consistently observe:
\begin{itemize}
\item A dominant tokenwise component that captures most of the full layer update,
\item A residual that is geometrically separated from this dominant tokenwise structure,
\item A strong association between approximation residual and output perturbation,
\item Structured layerwise regimes and a trade-off between expressivity and interpretability across function classes.
\end{itemize}

These results indicate that components not captured by tokenwise transformations are structured and systematically associated with changes in model outputs.

\section{Discussion}

Our results suggest that layer updates in deep language models are highly anisotropic. Across architectures, a large fraction of the layerwise update is captured by a restricted tokenwise transformation, while the remaining residual forms a geometrically distinct component. Empirically, the full update is strongly aligned with the tokenwise approximation, whereas the residual exhibits substantially weaker alignment, larger angular deviation, and lower projection onto the dominant tokenwise subspace. This indicates that the residual is not merely a small correction along the dominant update direction, but a qualitatively different component of the transformation.

This geometric separation has functional consequences. Approximation error under the restricted tokenwise model is strongly associated with output perturbation, showing that components outside the tokenwise class are disproportionately important for model behavior. In this sense, the residual provides a useful signal of where functionally significant computation occurs.

At the same time, the strength of this relationship depends on both regime and function class. In low-error regimes, simple linear maps often provide strong alignment between representation error and output perturbation. In higher-error regimes, this alignment weakens for linear maps and is partially recovered by more expressive models such as small MLPs. This suggests that the interpretability of the decomposition depends on whether the chosen function class is expressive enough to capture the dominant structure of the update, while still restrictive enough to leave a meaningful residual.

We do not interpret the residual as isolating a single mechanism. A nonzero residual may arise from cross-token interactions, nonlinear effects that are not representable within the chosen class, or other structured computations. Accordingly, we treat the residual as a function-class-dependent probe of transformation structure rather than a fixed architectural object. Components that appear as residual under one class may be absorbed by a more expressive one, so the notion of ``interaction'' in this framework is necessarily relative to the chosen approximation class.

This perspective is important because the decomposition is not inherited from architectural modules such as attention blocks or MLP blocks. Instead, it emerges from restricting the transition to tokenwise function classes, and therefore reflects a functional property of the learned representation dynamics rather than a structural property of the model design itself.

We also observe systematic architectural differences. Simpler function classes, such as diagonal transformations, are more effective in some models, while low-rank maps are more effective in others. This suggests that the tokenwise component of layer transitions has model-dependent geometry, and that different architectures may favor different low-complexity descriptions of their dominant update structure.

Our analysis also has clear limitations. The residual depends on the choice of function class, and expanding that class can reduce the residual magnitude. More expressive tokenwise models may improve approximation fidelity without necessarily preserving the same interpretable relationship between residual magnitude and output sensitivity. This highlights a distinction between minimizing approximation error and preserving a decomposition that remains structurally informative.

Taken together, our findings support a view in which most layerwise transformations behave like structured, approximately tokenwise updates, while functionally significant computation is concentrated in a geometrically distinct residual component. This provides a simple, architecture-agnostic lens for analyzing deep language models and suggests that identifying and characterizing such residual structure may be an important step toward understanding how meaningful computation is organized across layers.

A natural direction for future work is to further resolve the internal structure of the residual itself, for example by separating components that remain within the locally induced tokenwise subspace from those that lie outside it, or by identifying which kinds of context dependence are most consistently associated with the residual across models and layers.
\section*{Acknowledgements}
JY was supported by Brain Pool program funded by the Ministry of Science and ICT through the National Research Foundation of Korea (RS-2025-16322971).
\section*{Impact Statement}

This work introduces a simple empirical framework for analyzing layerwise transformations in deep sequence models through restricted function classes. By focusing on measurable approximation and functional impact, it complements existing interpretability approaches. While improved understanding of model internals may influence future system design and analysis, we do not identify specific risks requiring additional mitigation.

\bibliography{ref}
\bibliographystyle{icml2026}

\newpage
\appendix
\onecolumn

\section{Detailed Experimental Setup}
\label{app:experimental-setup}

This section provides complete experimental details for reproducibility.

\subsection{Models}

We evaluate our method across the following pretrained sequence models:
\begin{itemize}
    \item \textbf{DistilGPT2} (\texttt{distilgpt2})
    \item \textbf{Pythia-70M} (\texttt{EleutherAI/pythia-70m})
    \item \textbf{Pythia-410M} (\texttt{EleutherAI/pythia-410m})
    \item \textbf{Pythia-1B} (\texttt{EleutherAI/pythia-1b})
    \item \textbf{Pythia-1.4B} (\texttt{EleutherAI/pythia-1.4b})
    \item \textbf{Mamba-130M} (\texttt{state-spaces/mamba-130m})
    \item \textbf{Mamba-370M} (\texttt{state-spaces/mamba-370m})
\end{itemize}

All models are used in inference mode with publicly available pretrained checkpoints, without any fine-tuning or parameter updates.

\subsection{Dataset and Token Sampling}

We use the WikiText-103 validation split. Inputs are tokenized using the model-specific tokenizer and truncated or padded to a fixed sequence length of $192$ tokens.

From each input sequence, we uniformly sample token positions from valid (non-padding) positions. For each sampled position $t$, we collect pairs of hidden states from consecutive layers:
\[
(h_\ell, h_{\ell+1}).
\]

Across experiments, we use:
\begin{itemize}
    \item $2000$ training texts,
    \item $500$ test texts,
    \item $8$ sampled token positions per text,
\end{itemize}
resulting in approximately $1.7 \times 10^4$ token-level representation pairs per layer.

\subsection{Tokenwise Transformation Families}

For each layer $\ell$, we approximate the transition $h_\ell \rightarrow h_{\ell+1}$ using constrained tokenwise transformations that act independently on each token representation.

We consider the following transformation families:
\begin{itemize}
    \item \textbf{Local diagonal scaling (LOCAL\_DIAGPSD):} per-token diagonal linear maps with non-negative coefficients,
    \item \textbf{Global diagonal scaling (B1\_GLOBAL\_DIAGPSD):} a single diagonal transformation fitted globally,
    \item \textbf{Local low-rank linear maps (B2\_NN\_LR):} linear transformations constrained to rank $r \in \{4, 8, 16, 32, 64\}$,
    \item \textbf{Unitary transformations:} orthogonal linear maps preserving vector norms.
\end{itemize}

All transformations operate independently on each token, without cross-token interactions.

\subsection{Local Fitting Procedure}

For each reference representation $h_\ell^{(i)}$, we construct a local neighborhood $\mathcal{N}_k(i)$ using cosine similarity in representation space, with $k=64$ unless otherwise specified.

We then fit a local transformation $T$ by minimizing the least-squares objective:
\[
\sum_{j \in \mathcal{N}_k(i)} \| h_{\ell+1}^{(j)} - T(h_\ell^{(j)}) \|^2.
\]

Diagonal and low-rank transformations admit closed-form solutions. For low-rank maps, we compute the unconstrained least-squares solution and truncate its singular value decomposition to the specified rank.

\subsection{Intervention and Evaluation Metrics}

After fitting $T$, we perform an intervention by replacing the original transition with
\[
h_{\ell+1} \leftarrow T(h_\ell),
\]
and propagate the modified representation through the remaining layers to obtain a perturbed output distribution.

We evaluate the intervention using both functional and geometric metrics.

\paragraph{Relative Representation Error (RelErr).}
Normalized $\ell_2$ error between $T(h_\ell)$ and $h_{\ell+1}$.

\paragraph{Output Perturbation.}
KL divergence between the original and perturbed next-token distributions, averaged over tokens and samples.

\paragraph{Error--Perturbation Correlation.}
Spearman rank correlation $\rho$ between RelErr and output perturbation.

\paragraph{Directional Alignment.}
We define:
\[
\Delta_{\text{full}} = h_{\ell+1} - h_\ell, \quad
\Delta_{\text{tok}} = T(h_\ell) - h_\ell, \quad
r = h_{\ell+1} - T(h_\ell).
\]

We measure geometric alignment using absolute cosine similarity:
\[
\mathrm{Align}(v,u) = \frac{|v^\top u|}{\|v\|\|u\|},
\]
and report angular deviation in degrees.

We compute alignment between:
\begin{itemize}
    \item $\Delta_{\text{full}}$ and $\Delta_{\text{tok}}$,
    \item $r$ and $\Delta_{\text{tok}}$.
\end{itemize}

\paragraph{Subspace Projection.}
Let $U_k$ denote the top-$k$ left singular vectors of the fitted local map. For a vector $v$, we measure the fraction of energy captured by the dominant tokenwise subspace:
\[
\frac{\| U_k^\top v \|^2}{\|v\|^2}.
\]

We report this quantity for the full update, tokenwise update, and residual.

\paragraph{Signed Cosine (Auxiliary).}
We additionally report signed cosine similarity to detect directional bias, although all primary results use absolute cosine.

\subsection{Compute Details}

All experiments are implemented in PyTorch using HuggingFace Transformers. Experiments are run on an NVIDIA RTX 4060 GPU using FP32, except for larger models (Pythia-1B, 1.4B), where bf16 is used due to memory constraints.

\section{Sensitivity to Locality and Rank}

We evaluate the effect of neighborhood size $k$ and rank $r$ on approximation error.

\begin{figure}[h!]
    \centering
    \includegraphics[width=0.42\linewidth]{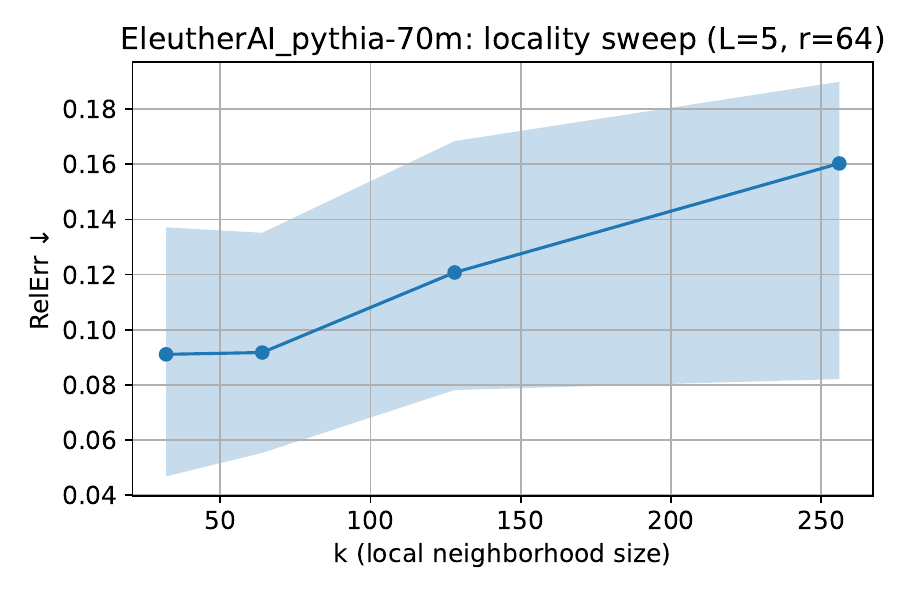}
    \includegraphics[width=0.42\linewidth]{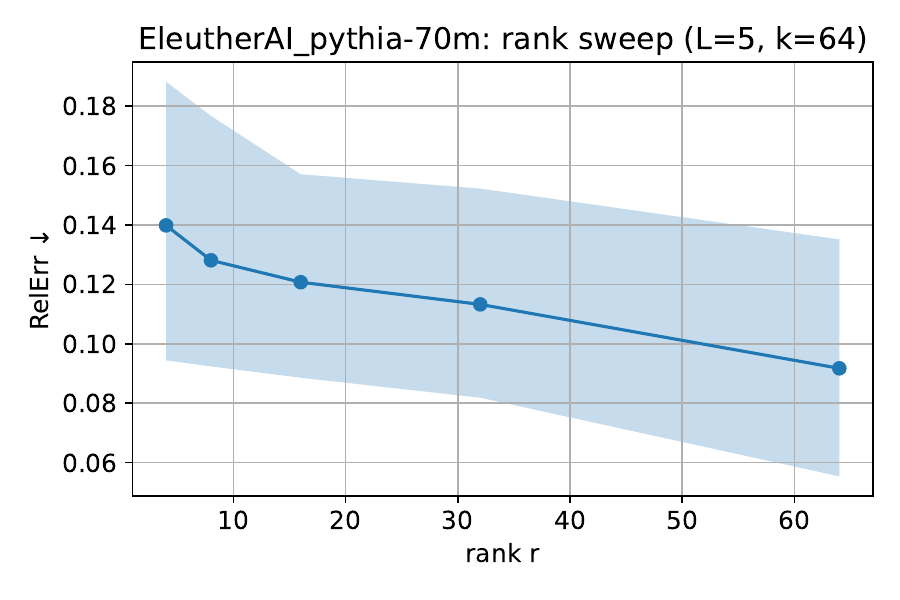}
    \caption{
    Sensitivity to locality and rank in Pythia-70M. 
    Left: neighborhood size sweep. 
    Right: rank sweep.
    }
    \label{fig:locality}
\end{figure}

Figure~\ref{fig:locality} shows that increasing $k$ improves stability of the local fit, while increasing rank $r$ reduces approximation error, indicating that the dominant tokenwise structure is low-dimensional but not strictly diagonal.

\end{document}